\documentclass[10pt,twocolumn,letterpaper]{article}

\usepackage{wacv}              

\usepackage{graphicx}
\usepackage{amsmath}
\usepackage{amssymb}
\usepackage{booktabs}
\usepackage{algorithm}
\usepackage{algpseudocode}
\usepackage{enumitem}
\usepackage{xcolor}
\usepackage{multirow}
\usepackage[accsupp]{axessibility} 

\usepackage[pagebackref,breaklinks,colorlinks]{hyperref}

\definecolor{blu}{rgb}{0.12,0.49,0.85}
\newcommand{\jh}[2]{\textcolor{black}{{}#2}}

\usepackage[capitalize]{cleveref}
\crefname{section}{Sec.}{Secs.}
\Crefname{section}{Section}{Sections}
\Crefname{table}{Table}{Tables}
\crefname{table}{Tab.}{Tabs.}

\def\wacvPaperID{259} 
\def\confName{WACV}
\def\confYear{2025}

\title{Tumor Synthesis conditioned on Radiomics}

\author{
    Jonghun Kim\textsuperscript{1}, \quad
    Inye Na\textsuperscript{1}, \quad
    Eun Sook Ko\textsuperscript{2}, \quad
    Hyunjin Park\textsuperscript{1 $\dagger$} \\
    \textsuperscript{1} Department of Electrical and Computer Engineering, 
    Sungkyunkwan University, Suwon, Korea \\
    \textsuperscript{2} Department of Radiology and Center for Imaging Science, Samsung Medical Center, \\
    Sungkyunkwan University School of Medicine, Suwon, Korea \\
    {\tt\small \{iproj2,niy0404,hyunjinp\}@skku.edu, mathilda0330@gmail.com}
}

\begin{document}
\twocolumn[{%
\renewcommand\twocolumn[1][]{#1}%
\maketitle
\begin{center}
    \vspace{-21pt}
    \centering
    \captionsetup{type=figure}
    \includegraphics[width=0.85\textwidth]{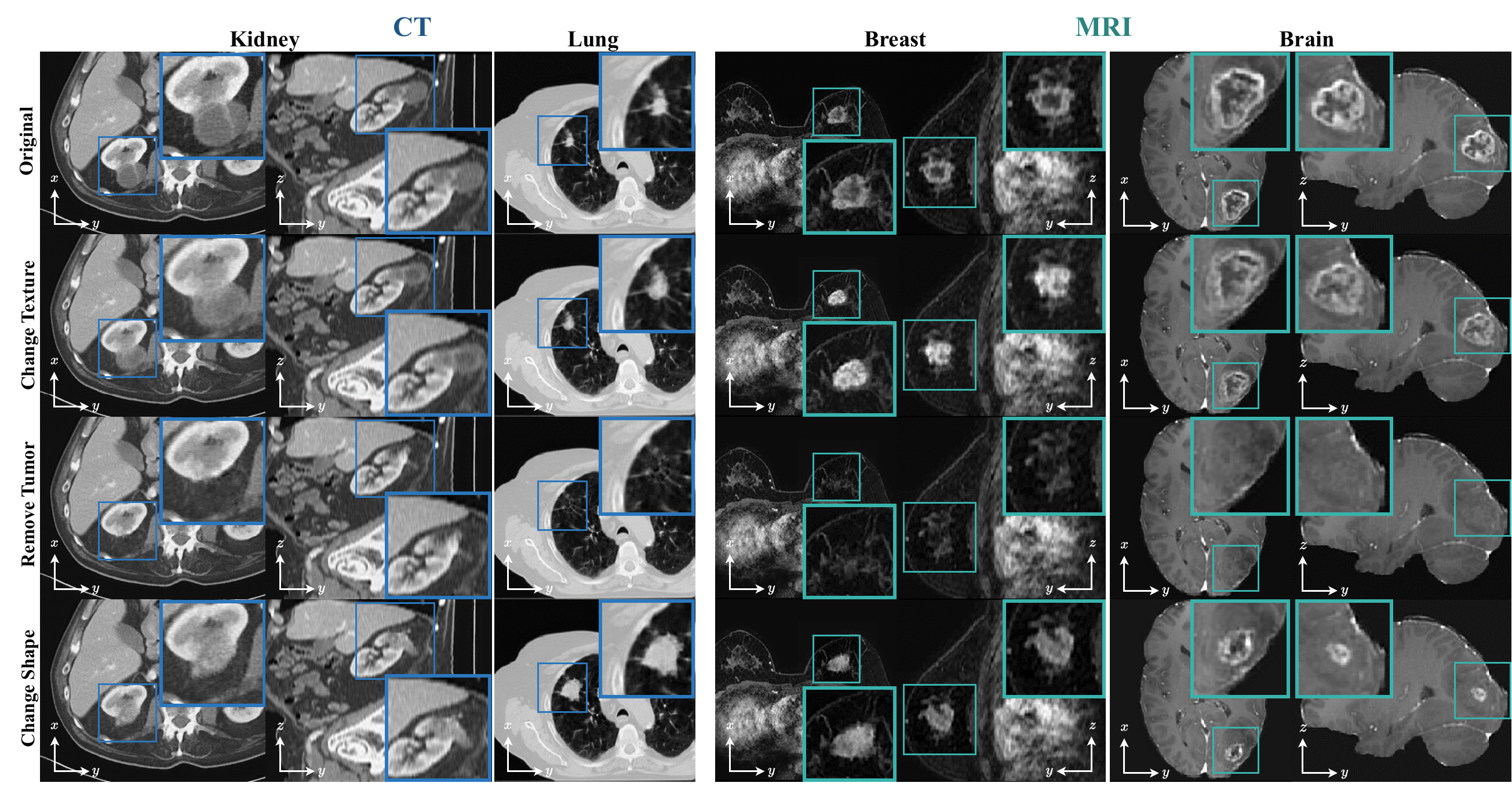}
    \vspace{-15pt}
    \captionof{figure}{Tumor synthesis results with proposed method. The image displays 2D planes of a volumetric image, with each plane labeled at the bottom left. The x-y plane illustrates the axial view, while the z-y plane depicts the sagittal view. First row: Original images, second row: Images with changed tumor texture, third row: Images with tumors removed, fourth row: Images with changed tumor shape and size.}
    \label{fig1}
    \vspace{6pt}
\end{center}%

}]

\begin{abstract}
\noindent 
Due to privacy concerns, obtaining large datasets is challenging in medical image analysis, especially with 3D modalities like Computed Tomography (CT) and Magnetic Resonance Imaging (MRI). Existing generative models, developed to address this issue, often face limitations in output diversity and thus cannot accurately represent 3D medical images. We propose a tumor-generation model that utilizes radiomics features as generative conditions. Radiomics features are high-dimensional handcrafted semantic features that are biologically well-grounded and thus are good candidates for conditioning. Our model employs a GAN-based model to generate tumor masks and a diffusion-based approach to generate tumor texture conditioned on radiomics features. Our method allows the user to generate tumor images according to user-specified radiomics features such as size, shape, and texture at an arbitrary location. This enables the physicians to easily visualize tumor images to better understand tumors according to changing radiomics features. Our approach allows for the removal, manipulation, and repositioning of tumors, generating various tumor types in different scenarios. The model has been tested on tumors in four different organs (kidney, lung, breast, and brain) across CT and MRI. The synthesized images are shown to effectively aid in training for downstream tasks and their authenticity was also evaluated through expert evaluations. Our method has potential usage in treatment planning with diverse synthesized tumors. Our code is available at \href{https://github.com/jongdory/TS-Radiomics}{github.com/jongdory/TS-Radiomics}.
\renewcommand{\thefootnote}{}
\footnotetext{$^\dagger$ Corresponding Author \\ This study was supported by National Research Foundation (RS-2024-00408040), Institute for Basic Science (IBS-R015-D1), AI Graduate School Support Program (Sungkyunkwan University) (RS-2019-II190421), ICT Creative Consilience program (RS-2020-II201821), and the Artificial Intelligence Innovation Hub program (RS-2021-II212068).}
\end{abstract}
    
\section{Introduction}
\label{sec:intro}

In medical image analysis, obtaining large datasets can be challenging due to privacy concerns \cite{shen2017deep, razzak2018deep}. To mitigate this, extensive research has been conducted on data augmentation based on generative models \cite{sandfort2019data, chen2022generative, na2023synthetic}. 
Training generative models, particularly those based on Generative Adversarial Networks (GANs) \cite{goodfellow2014generative}, faces significant challenges, including difficulty in model stability and interpretability of results in terms of their medical relevance. A notable issue with GANs is mode collapse, leading to a limited variety of outputs and reduced image diversity, which is particularly problematic in medical settings where accurate representation is vital. These challenges are further heightened with 3D medical images, requiring more extensive training samples for effective model training.


Early detection and prognosis diagnosis of tumors are important. Medical imaging, such as Computed Tomography (CT) and Magnetic Resonance Imaging (MRI), plays a pivotal role in the detection, diagnosis, staging, treatment response monitoring, and recurrence monitoring of tumors \cite{iyer2010mri, choi2014ct}. Radiologists can use medical imaging to understand a patient's condition and help select the best treatment method. However, visually evaluating tumors is subjective and can often miss subtle information due to the difficulty in recognizing fine textures or patterns \cite{lubner2017ct, bi2019artificial}.

Radiomics is the method of extracting hundreds to thousands of handcrafted semantic features from routine medical images, enabling quantitative analysis of subtle changes and complex patterns \cite{aerts2014decoding, gillies2016radiomics, lambin2017radiomics}. These features are based on the shape, pixel value distribution, texture, and other patterns of the region of interest. \jh{}{Radiomics extracts meaningful information in medical scenarios such as cancer diagnosis, prognosis, and treatment response prediction for various organs and scanners \cite{lambin2017radiomics, liu2019applications}. With proven medical efficacy, radiomics features are considered biologically well grounded  \cite{tomaszewski2021biological} and thus could be rich bases for tumor generation. This provides deep insights that are challenging to obtain through traditional manual analysis and can complement the judgments of clinicians \cite{lambin2017radiomics, hosny2018artificial, Kim_2024}.} However, some radiomics features such as complex texture are non-intuitive and challenge medical experts to grasp their significance.

We propose a tumor image generation model conditioned on radiomics features, leveraging the recent diffusion-based generative models \cite{ho2020denoising, rombach2022high, li2023bbdm}  and conditioning techniques through the cross-attention mechanism \cite{vaswani2017attention}. We demonstrate the ability to generate desired tumor images by adjusting low-dimensional radiomics features. This process involves manipulating intuitive radiomics features such as size to produce 3D tumor images. By converting radiomics features into images, we can provide visual insights. Furthermore, since our approach creates images through adjustable radiomics features, the rationale behind the outcomes for generation is clear. 
Our model facilitates the simulation of tumor characteristics, including location, size, shape, and texture in 3D medical imaging, enabling the creation of diverse and rare samples as needed. Validated through experiments on tumors in four different organs, it also allows for the generation of tumors with adjustable shapes and textures, as depicted in Figure \ref{fig1}. Our model might have future usage in treatment planning and prognosis prediction with diverse synthesized tumors leading to better personalized treatment options.

\noindent \textbf{Contribution}:
\begin{itemize}[noitemsep, topsep=0pt, partopsep=0pt, parsep=0pt]
    \item We suggest a tumor shape generator that uses a conditional GAN-based model using shape features and a tumor texture generator based on the Diffusion model to alter the texture of the tumor.
    \item We enable tumor synthesis through adjustable radiomics features. We propose a diffusion-based model capable of erasing tumors, changing their texture, and manipulating their shape, offering a more comprehensive approach to tumor analysis and simulation.
    \item We have validated our generative model on tumors across four different organs using CT and MRI images, demonstrating our model's effectiveness through visual results.
    \item The synthesized images have been useful in aiding downstream tasks and deemed realistic according to expert evaluations.
\end{itemize}    

\section{Related Work and Backgrounds}
\label{sec:related_work}

\begin{figure*} [t]
    \centering
    \vspace{-6pt}
    \includegraphics[width=1\textwidth]{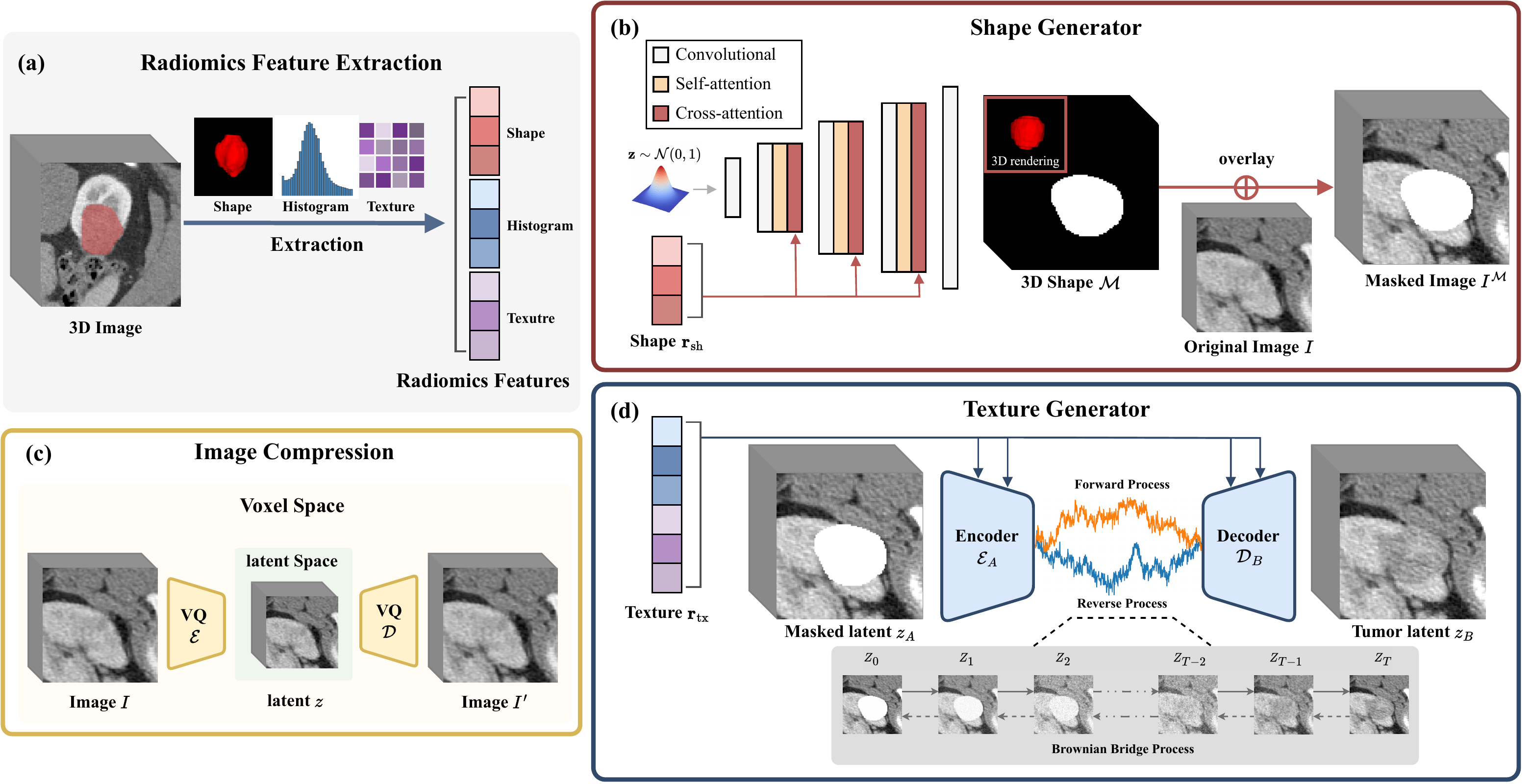}
    \vspace{-12pt}
    \caption{The pipeline of the proposed method. (a) Radiomics feature extraction illustrates the extraction of shape, histogram, and texture features from 3D image and tumor mask. (b) Shape generator depicts training a GAN-based model conditioned on shape features to generate a tumor mask. (c) Image compression into latents using VQ-GAN. (d) Texture generator demonstrates overlaying the generated mask from (b) onto a 3D image to create a masked latent and training a diffusion-based model conditioned on texture features to generate a tumor latent, illustrating the transition from masked domain A to tumor domain B.}
    \label{fig2}
    \vspace{-12pt}
\end{figure*}

\noindent \textbf{Generative Adversarial Networks} are generative models with a structure where a generator and a discriminator learn through competition \cite{goodfellow2014generative, karras2019style}. With the enhancement in sampling quality and diversity of GANs, they have been deployed across various computer vision applications such as text-to-image synthesis \cite{reed2016generative, zhang2017stackgan, xu2018attngan, tao2022df}, image-to-image translation \cite{isola2017image, huang2018multimodal, yu2019ea, kong2021breaking, dalmaz2022resvit}, and image editing \cite{zhu2016generative, patashnik2021styleclip}. They have rapidly advanced image generation and led to various applications across different domains. 
\jh{}{They have also been studied in medical image analysis to solve various problems. \cite{yu2019ea, kong2021breaking, dalmaz2022resvit}. For instance, GANs have been utilized for image translation, converting MRI to CT or PET images \cite{dong2019synthetic, ben2019cross}, or facilitating modality transitions within MRIs \cite{yu2019ea, kong2021breaking, dalmaz2022resvit}. Moreover, they have been employed in denoising low-dose CT images to enhance the quality \cite{li2021low}. Given the typical scarcity of data in medical image domains, traditional training  can be challenging. However, research using GANs to augment data in medical imaging has shown that training with the generated data can improve performance \cite{sandfort2019data, chen2022generative}. Additionally, there are studies on synthesizing tumors in 2D images using radiomics features in GANs \cite{na2023synthetic}.} Nonetheless, GANs can have unstable learning phases and issues like mode collapse, leading to generating specific data only \cite{saxena2021generative, srivastava2017veegan}.

\vspace{3pt}
\noindent \textbf{Diffusion probabilistic models (DPMs)} \cite{sohl2015deep, ho2020denoising, song2021scorebased} aim to learn the diffusion process that generates the probability distribution of a given dataset. Unlike GANs, DPMs are known for converging well even with fewer hyperparameters and for producing sharp and detailed images \cite{yang2022diffusion}. They have been applied in various domains including text-to-image \cite{rombach2022high, ramesh2022hierarchical, zhang2023adding}, image-to-image \cite{saharia2022palette, li2023bbdm, zhang2023adding, Kim_2024_WACV}, text-to-video \cite{singer2023makeavideo, wu2023tune}, data augmentation \cite{feng2023diverse, trabucco2023effective}, super-resolution \cite{ho2022cascaded, saharia2022image, yue2023resshift}, image inpainting and outpainting \cite{rombach2022high, saharia2022palette, lugmayr2022repaint, zhang2023adding}. 
\jh{}{Additionally, they have been utilized in the medical domain for various task-specific tasks such as generation, segmentation, translation, anomaly detection, and registration\cite{na2024radiomicsfill, wu2024medsegdiff, kim2022diffusemorph, pinaya2022brain, wyatt2022anoddpm, Kim_2024_WACV, kazerouni2023diffusion}.} The Latent Diffusion Model (LDM) \cite{rombach2022high} uses the latent space for high-resolution image generation with efficient computation compared to traditional diffusion models. It utilizes the Vector Quantized GAN (VQGAN) \cite{esser2021taming} to effectively quantize and compress the latent space, aiding in representing and manipulating image features. LDM has shown potential in various image synthesis tasks, demonstrating its ability to handle diverse generation tasks efficiently \cite{rombach2022high, pinaya2022brain, blattmann2023align, jiang2023cola}.

\noindent \textbf{Brownian Bridge Diffusion Model (BBDM)} assumes the diffusion process as a probabilistic Brownian bridge process in image-to-image translation tasks \cite{li2023bbdm}. By constructing a direct mapping between the source and target domains, it provides a potentially more efficient and generalized model for image-to-image translation tasks, showcasing its applicability across a range of domains. In BBDM, Given the source domain $\boldsymbol{x}_0$ and target domain $\boldsymbol{y}$, the forward process is defined as:
\begin{equation}  
q(\boldsymbol{x}_t | \boldsymbol{x}_{0}, \boldsymbol{y}) = \mathcal{N}(\boldsymbol{x}_t;(1-m_t)\boldsymbol{x}_{0} + m_t\boldsymbol{y}, \delta_t\boldsymbol{I})
\label{eq:1}
\end{equation}
\noindent , where $m_t = \frac{t}{T}$ with T representing the total steps of the diffusion process and $\delta_t$ is a fixed variance. The intermediate state $\boldsymbol{x}_t$ is defined in a discrete form as follows:
\begin{equation}  
\boldsymbol{x}_t = (1-m_t)\boldsymbol{x}_0 + m_t\boldsymbol{y} + \sqrt{\delta_t}\boldsymbol{\epsilon}_t
\label{eq:2}
\end{equation}
\noindent, where $\epsilon_t \sim \mathcal{N}(\boldsymbol{0},\boldsymbol{I})$ is the Gaussian noise. Then, BBDM is trained to approximate the reverse process: 
\begin{equation}  
p_{\theta}(\boldsymbol{x}_{t-1} | \boldsymbol{x}_t, y) = \mathcal{N}(\boldsymbol{x}_{t-1};\boldsymbol{\mu}_\theta(\boldsymbol{x}_t,\boldsymbol{y},t), \tilde{\delta}_t\boldsymbol{I})
\label{eq:3}
\end{equation}
\noindent, where $\boldsymbol{\mu}_\theta(\boldsymbol{x}_t,t)$ represents the predicted mean value of the noise and $\tilde{\delta}_t$ denotes the variance of noise at each step. $\boldsymbol{\mu}_t$ is a mean parameterized by the noise predictor $\boldsymbol{\epsilon}_\theta$:
\begin{equation}  
\boldsymbol{\mu}_{\theta}(\boldsymbol{x}_t, \boldsymbol{y}, t) = c_{xt}\boldsymbol{x}_t + c_{yt}\boldsymbol{y} + c_{\epsilon t}\boldsymbol{\epsilon}_{\theta}(\boldsymbol{x}_t, t)
\label{eq:4}
\end{equation}
, where $c_{xt},  c_{yt},  c_{\epsilon t}$ are constants varying with respect to time step $t$. In the translation process, the sample can be obtained from the Gaussian noise by iterative reverse process: $\boldsymbol{x}_{t-1}=\mu_\theta(\boldsymbol{x}_t,\boldsymbol{y},t) + \sqrt{\tilde{\delta}_t}\boldsymbol{z}$, where $\boldsymbol{z} \sim \mathcal{N}(0,1)$
In this study, we utilize radiomics texture features as conditioning in the BBDM to perform the translation task from tumor-masked images to tumor images.

\section{Method}
\label{sec:method}

\setlength{\abovedisplayskip}{5pt}
\setlength{\belowdisplayskip}{5pt}

We generate tumor images by utilizing radiomics features extracted using the tumor mask and the underlying 3D medical images. 
\jh{}{The entire pipeline of our proposed method is depicted in Figure \ref{fig2}. To effectively synthesize tumors, we generate the shape and texture separately. The shape generator and texture generator are trained independently. First, the shape is generated and masked onto the desired image location, then the texture is generated using radiomics features. For normal tissue generation, the tumor area is masked out, and the texture is generated without using radiomics features. Details on the shape generator are in Section \ref{sec:3.2}, and the texture generator in Section \ref{sec:3.3}.}

\begin{figure} [t]
    \vspace{-9pt}
    \centering
    \includegraphics[width=\columnwidth]{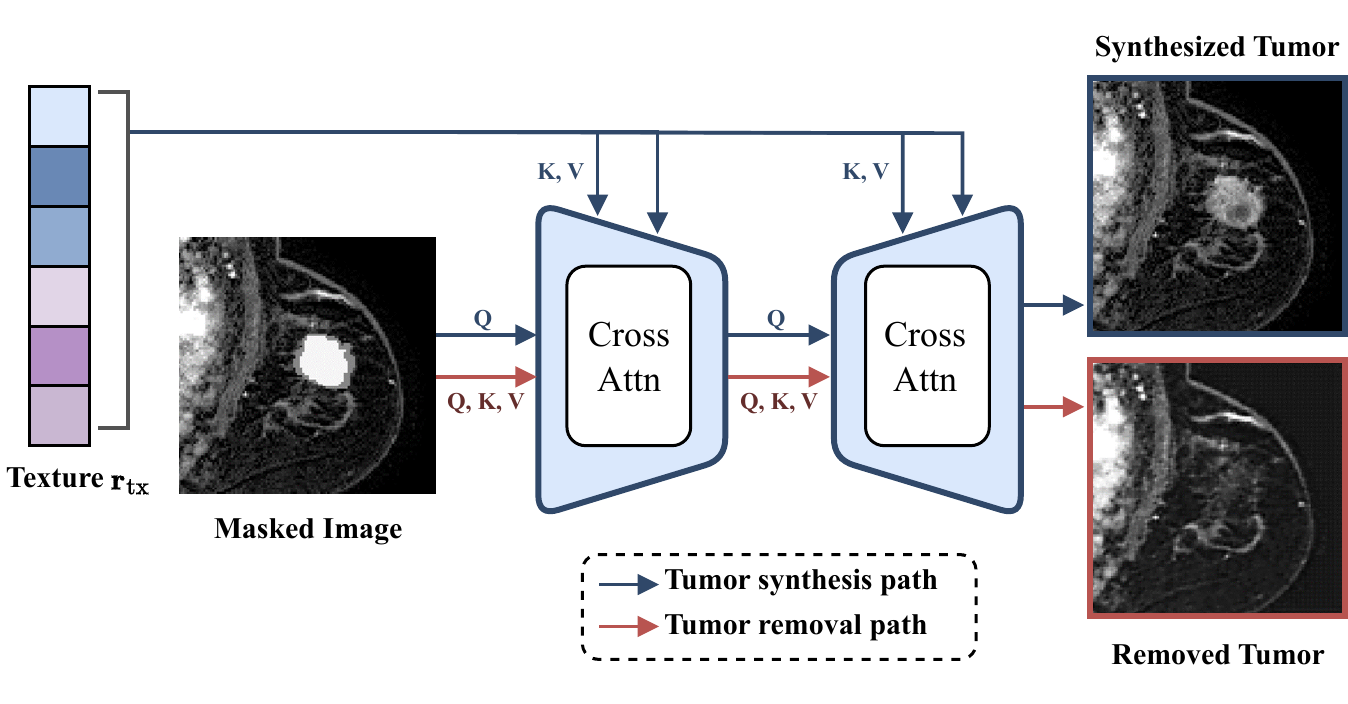}
    \vspace{-21pt}
    \caption{Two generation paths within the texture generator. Blue path: the tumor synthesis path, where texture features are used as key and value for cross-attention, generating the synthesized tumor image. Red path: tumor removal path, where self-attention is performed to generate the removed tumor image.}
    \label{fig3}
    \vspace{-9pt}
\end{figure}

\subsection{Tumor Shape Generator}
\label{sec:3.2}
\jh{}{We employed a GigaGAN \cite{kang2023scaling} with cross-attention \cite{vaswani2017attention} for our shape generator, adapting it to a 3D method. We use radiomics features rather than text for feature-to-image generation, focusing on shape-feature conditioning. Since shape masks are relatively easy to generate, we used a simple GAN instead of a complex diffusion model. Traditional convolution filters are limited to their receptive fields \cite{alzubaidi2021review} and this limitation is significant in the context of tumor shape, where features like volume, surface area, sphericity, and diameter are influenced by long-range relationships. Therefore, integrating these relationships using attention layers is essential. We utilize self-attention to assimilate long-range relationships and cross-attention to enable the generation of tumor shapes with texture features, as depicted in Figure \ref{fig2} (b).} Our shape generator $G$, generates the shape $\mathcal{M}$, in conjunction with a latent $\mathbf{z} \sim \mathcal{N}(0,1)$, and shape feature $\mathbf{r}_{\mathrm{sh}}$.
\begin{equation}  
\mathcal{M} = G(\mathbf{z}, \mathbf{r}_{\mathrm{sh}})
\label{eq:5}
\end{equation}
\noindent, where shape is  $\mathcal{M} \in \mathbb{R}^{H\times W\times D}$ and shape feature is $\mathbf{r}_{\mathrm{sh}} \in \mathbb{R}^{d_{\mathrm{sh}}}$.
We can obtain a masked image $I^{\mathcal{M}}$ by applying a mask at the desired location for tumor generation using shape $\mathcal{M}$.

\begin{figure*} [t]    
        \vspace{-12pt}
        \centering
        \includegraphics[width=0.95\textwidth]{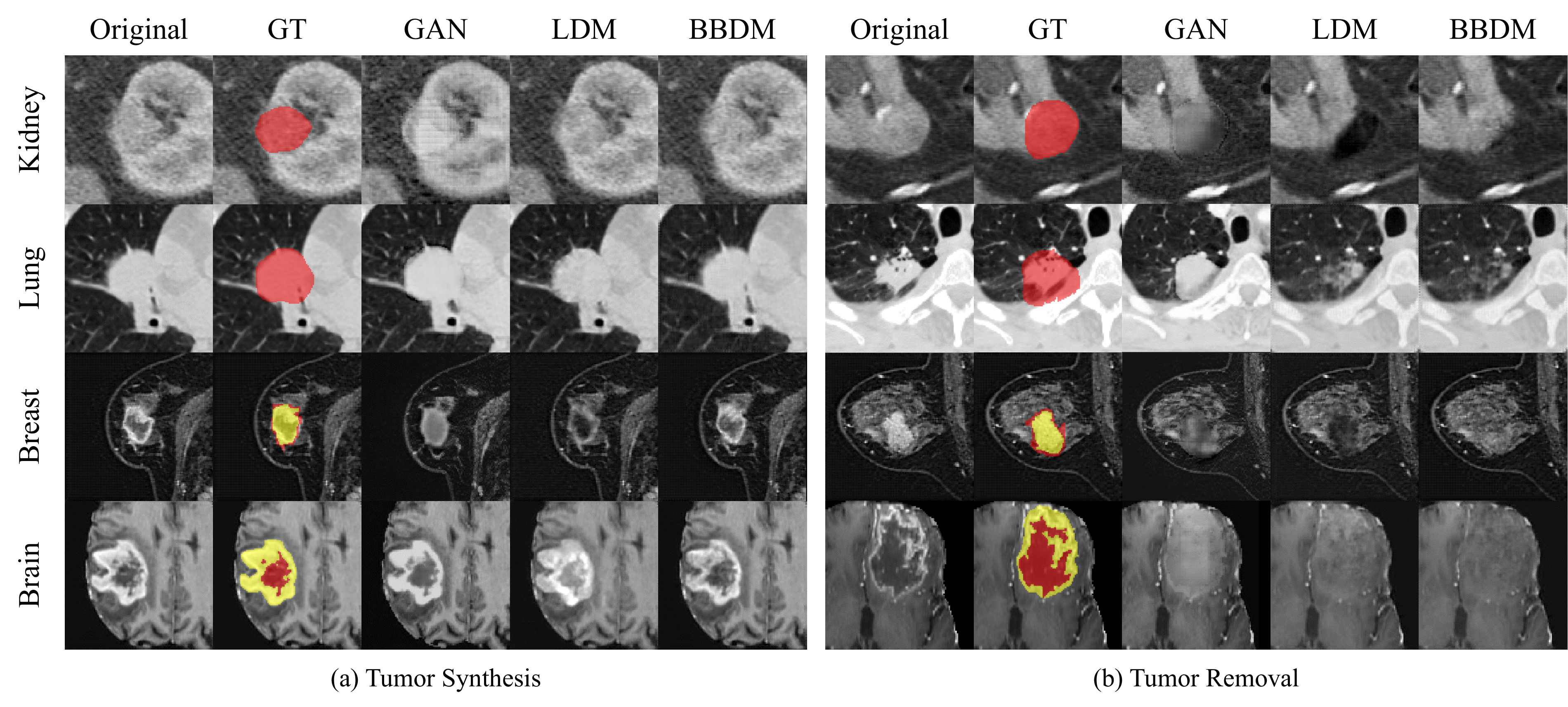}
        \vspace{-9pt}
        \caption{Qualitative results of the Baseline on four organs. This displays the outcomes of each model performing two tasks. Synthesis:  presents the tumor synthesis result in the given mask when provided with radiomics texture features. Removal: shows the result after removing the tumor corresponding to the mask. In breast, red: peri-tumor, yellow: tumor. In brain, red: necrosis, yellow: enhancing tumor. }
        \label{fig4}
        \vspace{-9pt}

\end{figure*}

\subsection{Tumor Texture Generator}
\label{sec:3.3}
\vspace{-12pt}
\jh{}{In contrast to the shape generator, to create complex tumor texture patterns in 3D medical images like CT and MRI, we use BBDM \cite{li2023bbdm}, which applies a Brownian Bridge process to a diffusion model. This generates texture from the masked image to synthesize the tumor image. Unlike DDPM, which generates the target image from noise, BBDM is specialized for Image-to-Image translation, generating the target domain image from a source domain image.  We employ this to translate from the masked domain A to the tumor domain B, as illustrated in Figure \ref{fig2} (d).} Due to the large dimensions of the original 3D volume, we use image compression with VQ-GAN \cite{esser2021taming, rombach2022high} and then conducted the diffusion process in the latent space, as shown in Figure \ref{fig2} (c). Similar to previous studies \cite{ho2020denoising, li2023bbdm}, we utilize time conditioning UNet \cite{ronneberger2015u, ho2020denoising} $\epsilon_\theta$ as the backbone. We enable the generation of tumor images corresponding to the given texture condition by utilizing texture features. We follow the standard training object for BBDM and use texture feature $\mathbf{r}_\mathrm{tx}$ as conditioning:
\begin{equation}
\mathbb{E}_{\boldsymbol{x}_0, \boldsymbol{y}, \boldsymbol{\epsilon}}[c_{\epsilon t} || m_t(\boldsymbol{y} - \boldsymbol{x}_0) + \sqrt{\delta_t}\boldsymbol{\epsilon} - \boldsymbol{\epsilon}_\theta(\boldsymbol{x}_t,t,\mathbf{r}_{\mathrm{tx}})||^2]
\label{eq:6}.
\end{equation}
\jh{}{To allow more flexibility in adjusting tumors, we train both tumor synthesis and tumor removal simultaneously. This method uses self-attention when texture features are not provided as conditions and switches to cross-attention when texture conditions are provided. This approach not only helps the model learn the characteristics of surrounding the organ but also eliminates the need to train separate models for tumor generation and removal.}
This process is illustrated in Figure \ref{fig3}. Similar to previous studies \cite{rombach2022high}, the conditioning mechanisms through cross-attention in the intermediate layers of the UNet are defined as follows:
Attention($Q,K,V$) = softmax($\frac{QK^T}{\sqrt{d}}\cdot V$) with 
\begin{equation}
Q = W_{Q}^{(i)} \cdot q, K = W_{K}^{(i)} \cdot k, V = W_{V}^{(i)} \cdot v
\label{eq:7}
\end{equation}
, where $W_{Q}^{(i)} \in \mathbb{R}^{d\times d_q^i}, W_{K}^{(i)} \in \mathbb{R}^{d\times d_k^i}$ and $W_{V}^{(i)} \in \mathbb{R}^{d\times d_v^i}$ are learnable projection matrices. $q, k, v$ depend on the existence of an input texture feature $\mathbf{r}_\mathrm{tx}$:
\begin{equation}
\begin{cases} 
q = \varphi_{i}(z_{t}),\quad k,v= \mathbf{r}_\mathrm{tx} &\quad  \text{if $\mathbf{r}_\mathrm{tx}$ is given,} \\
q,k,v = \varphi_{i}(z_{t}) &\quad  \text{otherwise}
\end{cases}
\label{eq:8}
\end{equation}
, where $\varphi_{i}(z_{t}) \in \mathbb{R}^{N\times d_{\epsilon}^i}$ denotes an intermediate representation of the UNet implementing $\epsilon_{\theta}$. Therefore, when texture features are not provided, we can reconstruct the masked area and by utilizing this, we can also remove the tumor, allowing for flexible alteration of the tumor shape. When training the tumor removal path, it creates a masked image by randomly applying a mask to an area that does not overlap with the existing tumor location. \jh{}{During inference for tumor removal, the tumor area is masked by the user-provided mask and then removed via the removal path. For tumor synthesis, the normal area is masked by the synthesized mask from the shape generator and texture features are provided as input to synthesize the tumor through the synthesis path.} For detailed training and inference processes, please refer to Algorithms \ref{algorithm1}, \ref{algorithm2} and \ref{algorithm3} of supplementary.

\section{Experiments}
\label{sec:experiments}

\begin{table*} [t]
    \begin{minipage} {1\textwidth}
        \vspace{-12pt}
        \captionsetup{type=table}
        \centering
        \scalebox{0.75}{
        \begin{tabular}{lcccccccccccc}
            \toprule
             Task &  \multicolumn{6}{c}{(a) Tumor Synthesis} &  \multicolumn{6}{c}{(b) Tumor Removal} \\
             \cmidrule(lr){2-7} \cmidrule(lr){8-13}
             Model &  \multicolumn{2}{c}{GAN} &  \multicolumn{2}{c}{LDM} &  \multicolumn{2}{c}{\textbf{Ours}} &  \multicolumn{2}{c}{GAN} &  \multicolumn{2}{c}{LDM} & \multicolumn{2}{c}{\textbf{Ours}} \\
             \cmidrule(lr){2-3} \cmidrule(lr){4-5} \cmidrule(lr){6-7} \cmidrule(lr){8-9} \cmidrule(lr){10-11} \cmidrule(lr){12-13} 
            Organ (Modality) \hspace{0.05cm} & PSNR $\uparrow$ & SSIM $\uparrow$ & PSNR $\uparrow$ & SSIM $\uparrow$ & PSNR $\uparrow$ & SSIM $\uparrow$ & PSNR $\uparrow$ & SSIM $\uparrow$ & PSNR $\uparrow$ & SSIM $\uparrow$ & PSNR $\uparrow$ & SSIM $\uparrow$  \\
            \cmidrule(lr){1-13}
            Kidney (CT)  & 25.62 & 0.6962 & 30.57 & 0.8829 & \textbf{33.95}$^*$ & \textbf{0.9346}$^*$ & 25.88 & 0.6985 & 31.02 & 0.8877 & \textbf{34.16}$^*$ & \textbf{0.9370}$^*$ \\
            Lung (CT)    & 23.10 & 0.6628 & 29.35 & 0.9287 & \textbf{32.68}$^*$ & \textbf{0.9471}$^*$ & 23.61 & 0.6704 & 28.42 & 0.9205 & \textbf{32.81}$^*$ & \textbf{0.9484}$^*$ \\
            Breast (MRI) & 23.44 & 0.6420 & 29.51 & 0.8726 & \textbf{31.85}$^*$ & \textbf{0.9177}$^*$ & 24.04 & 0.6516 & 29.86 & 0.8741 & \textbf{32.07}$^*$ & \textbf{0.9193}$^*$ \\
            Brain (MRI)  & 30.17 & 0.9038 & 35.83 & 0.9616 & \textbf{36.67}$^*$ & \textbf{0.9634}$^*$ & 29.59 & 0.8943 & 35.91 & 0.9609 & \textbf{37.24}$^*$ & \textbf{0.9681}$^*$ \\
            \bottomrule
        \end{tabular}
        }
        \vspace{-6pt}
        \begin{flushright}
            \scalebox{1}{
                \scriptsize{$*p$-value$\ < 0.05$ comparing two best results}
            }
        \end{flushright}
        \vspace{-12pt}
        \caption{Quantitative comparison results for four organs and two modalities were measured using PSNR and SSIM.} 
        \label{table1}
    \end{minipage}
\end{table*}

\noindent \textbf{Datasets.}
To substantiate the effectiveness of our method, we validated tumor synthesis performance across two modalities and four organs. The kidneys and lungs were examined using CT data with the kidney data derived from the KiTS23 \cite{heller2023kits21} and the lung data from the NSCLS \cite{aerts2015data}. For the breast and brain, MRI data were utilized with the breast data coming from a private dataset and the brain data from the BraTS2021 \cite{baid2021rsna, menze2014multimodal, bakas2017advancing}. We allocated 80\% of all datasets for training and 20\% for testing. Details of the datasets are available in the \cref{sec:A_datasets} of supplementary.

\vspace{3pt}
\noindent \textbf{Models and hyperparameters.}
Our tumor shape generator is built upon GigaGAN \cite{kang2023scaling}, while the tumor texture generator is based on BBDM \cite{li2023bbdm}. Due to the large dimensions of 3D volumes and the consequent high computational cost, we employ VQGAN \cite{esser2021taming} to compress and reduce the model size. BBDM consists of two components: a pretrained VQGAN and a Brownian Bridge diffusion model. The VQGAN is later utilized as a comparative model, employing the same model as the Latent Diffusion Model used in subsequent comparisons. During the training stage, the number of time steps for the Brownian Bridge was set to 1000, while in the inference stage, 200 sampling steps were used. The implementation was done using PyTorch\footnote{\url{https://pytorch.org/}} and MONAI\footnote{\url{https://monai.io/}} libraries. We train the network by using the Adam \cite{kingma2014adam} optimizer with a learning rate of $5\times10^{-6}$. Training was conducted on four A100 80GB GPUs with a batch size of 1 per GPU. For details on the model architecture and hyperparameters, refer to the \cref{sec:B_implementation} of supplementary.

\vspace{1pt}
\noindent \textbf{Baseline Methods and Metrics.}
\jh{}{To our knowledge, there has not been much previous research that used radiomics features to create tumors for 3D images. Therefore, to substantiate the efficiency of our tumor texture generator, we compare ours with two baselines: one based on GAN and the other on LDM. The GAN-based model is adapted by modifying the 3D image translation model Ea-GAN \cite{yu2019ea} (based on pix2pix \cite{isola2017image}), with the addition of cross-attention for conditioning radiomics.} All the baselines we used in the experiment, as well as our method, are 3D methods using radiomics conditioning. \jh{}{For tumor generation, all texture generators used the same shape generator, and in the case of LDM, the same image compression method as BBDM was used.} We employed Peak Signal-to-Noise Ratio (PSNR) and the Structural Similarity Index Measure (SSIM), commonly used quantitative evaluations in medical image generation, to assess the quality of image synthesis \cite{yi2019generative}. Since the radiomics features can significantly change based on the preprocessing and normalization of the given modality, we assessed whether the features of generated images match the original radiomics features given as conditions by measuring the Pearson and Spearman correlation coefficient.

\section{Results}

\noindent \textbf{Results of Comparison Method}: To demonstrate the effectiveness of our method, we compared its generative performance on a test set against that of the baselines. We conducted both qualitative and quantitative evaluations. The results of the qualitative assessment are depicted in Figure \ref{fig4}. Each model performed tumor synthesis and tumor removal tasks simultaneously through cross-attention mechanisms. GAN-based model struggled to produce natural depictions for both tasks. This is largely due to the difficulty of obtaining ample data in the medical imaging field and the challenge of dealing with 3D images, which prevented the GAN-based model from generating high-quality images in both tasks across organs. However, models based on diffusion processes could generate good-quality images that closely resembled the real images. Additionally, the BBDM performed the tumor removal task quite naturally, especially in comparison to the LDM. The quantitative evaluation results for each baseline are in Table \ref{table1}. Consistent with the qualitative assessment, the BBDM exhibits the best performance in generating images for all organs.

\begin{table}[t]
    \vspace{-3pt}
    \centering
    \begin{minipage}{0.46\textwidth}
        \centering
        \scalebox{0.75}{
            \begin{tabular}{lcccccc}
                \toprule
                 Model &  \multicolumn{2}{c}{GAN} &  \multicolumn{2}{c}{LDM} &  \multicolumn{2}{c}{\textbf{Ours}} \\
                \cmidrule(lr){2-3} \cmidrule(lr){4-5} \cmidrule(lr){6-7}
                Organ \hspace{0.5cm} & PCC $\uparrow$ & SCC $\uparrow$ & PCC $\uparrow$ & SCC $\uparrow$ & PCC $\uparrow$ & SCC $\uparrow$   \\
                \cmidrule(lr){1-7}
                Kidney & 0.553 & 0.587 & 0.793 & 0.818 & \textbf{0.829}$^*$ & \textbf{0.865}$^*$ \\
                Lung   & 0.591 & 0.624 & 0.759 & 0.827 & \textbf{0.801}$^*$ & \textbf{0.859}$^*$ \\
                Breast & 0.610 & 0.657 & 0.828 & 0.810 & \textbf{0.852}$^*$ & \textbf{0.851}$^*$ \\
                Brain  & 0.645 & 0.672 & 0.849 & 0.883 & \textbf{0.864}$^*$ & \textbf{0.906}$^*$ \\
                \bottomrule
            \end{tabular}
        }
        \vspace{-6pt}
        \begin{flushright}
            \scalebox{1}{
                \scriptsize{$*p$-value$\ < 0.05$ comparing two best results}
            }
        \end{flushright}
        \centering
        \vspace{-12pt}
        \caption{Correlation between conditioned texture features $\mathbf{r}_{tx}$ and the texture features extracted from generated images.}
        \label{table2}
    \end{minipage}
\end{table}

\begin{table}[t]
    \centering
    \begin{minipage}{0.46\textwidth}
        \scalebox{0.7}{
            \begin{tabular}{ll|cccc|c}
                \toprule
                Metric & Model &  Kidney & Lung & Breast & Brain & Avg.\\
                \midrule 
                \multirow{2}{*}{Pearson Corr.} & GAN  &  0.937 & 0.928 & 0.893 & 0.902 & 0.915\\
                & BBDM  &  0.925 & 0.933 & 0.913 & 0.905 & 0.919 \\
                \midrule
                \multirow{2}{*}{Spearman Corr.} & GAN  & 0.954 & 0.941 & 0.907 & 0.916 & 0.930\\
                & BBDM  &  0.932 & 0.939 & 0.924 & 0.921 & 0.929 \\
                \bottomrule
            \end{tabular}
            }
        \centering
        \vspace{-6pt}
        \caption{Correlation between conditioned shape features $\textbf{r}_{sh}$ and the shape features extracted from generated tumor masks.}
        \label{table3}
        \vspace{-6pt}
    \end{minipage}
\end{table}

We also evaluated how well the generated images reflect the given radiomics features used as conditions. We re-extracted the radiomics features from the generated images and conducted a correlation comparison with the radiomics features provided as conditions. Table \ref{table2} shows the correlations between the original texture feature and the extracted texture features from the images generated by each baseline. We observed that the fidelity in terms of correlation of reproducing radiomics features was relatively higher in MRI images, with the highest fidelity occurring in the brain. Moreover, the degree of reproduction was somewhat lower in CT images compared to MRI, possibly because tumor regions in CT are generally more ambiguous.

\begin{figure} [t]
    \vspace{-9pt}
    \begin{minipage}{0.46\textwidth}
        \centering
        \includegraphics[width=0.8\columnwidth]{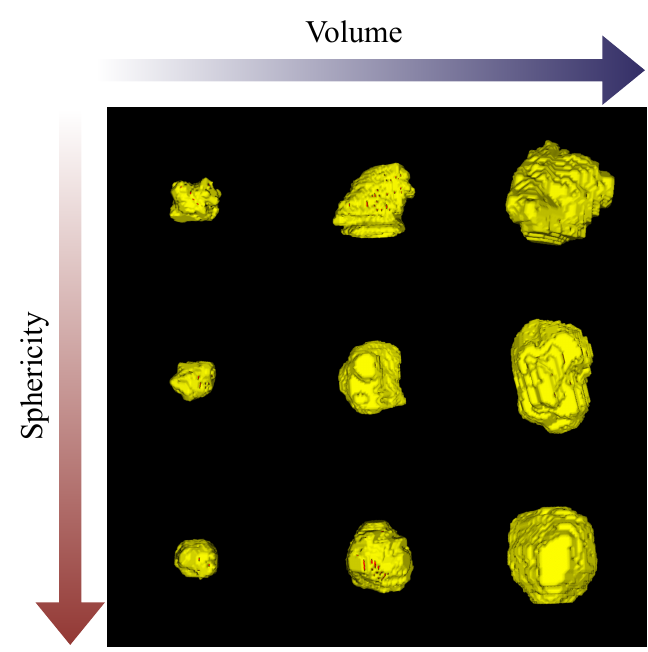}
        \captionsetup{type=figure}
        \caption{The results of the Tumor Shape Generator for breast. The tumors are conditioned by shape features. Volume denotes the size of the tumor, increasing from left to right. Sphericity refers to the degree to which the shape approaches that of a sphere, assessed from top to bottom.}
        \label{fig5}
        \vspace{-9pt}
    \end{minipage}
\end{figure}

\noindent \textbf{Manipulation of Tumor Shape}:
We generated 3D tumor masks with the appropriate conditions using shape features in the tumor shape generator. We quantitatively assessed the generated tumor masks by comparing the shape feature values of the ground truth with those of the generated tumor masks using correlations. \jh{}{Table \ref{table3} shows the correlations between the original shape feature and the extracted shape features from the tumor mask generated by the shape generator. This demonstrates that the generation process accurately integrated the shape features showing that there is little difference in performance between GAN and BBDM for shape generation. It also proves that even a simple GAN is sufficient for effective modeling.} The shape features used are specified in \cref{sec:C_radiomics} of the supplementary.

\begin{figure*}[t]
    \centering
    \begin{minipage}{\textwidth}
        \centering
        \vspace{-15pt}
        \includegraphics[width=0.825\textwidth]{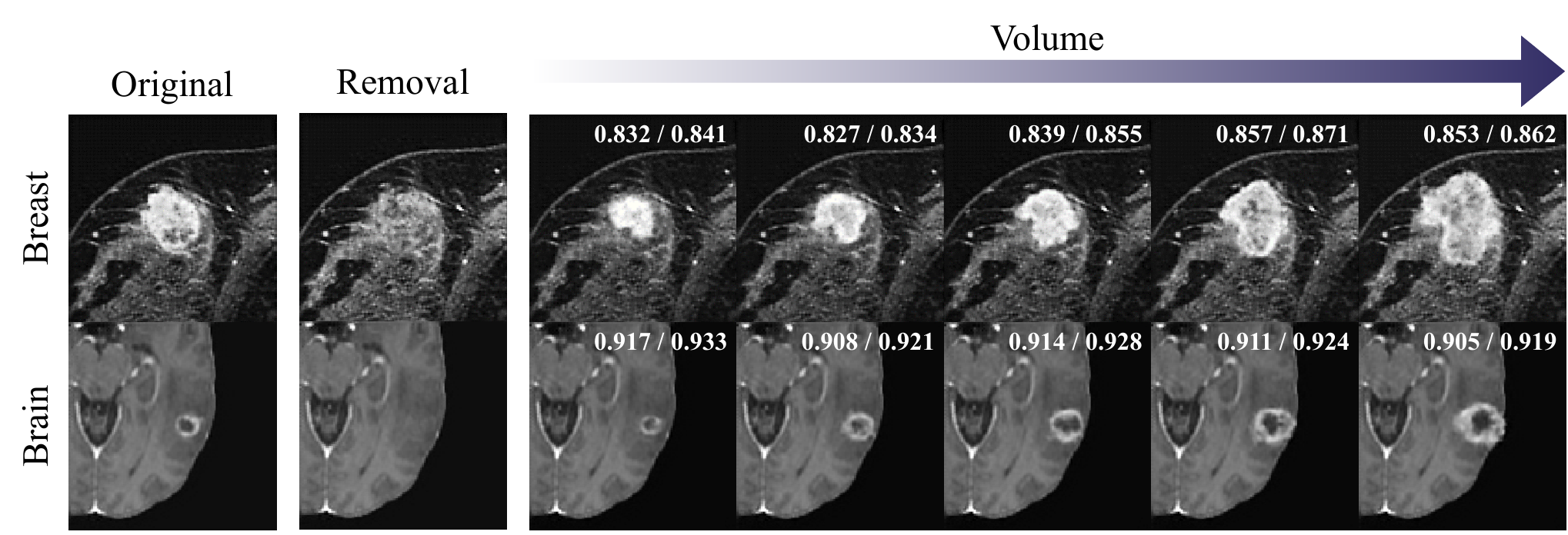}
        \vspace{-3pt}
        \caption{Results of tumor generation for various volume sizes using a tumor mask created by the tumor shape generator according to specified volume size features. The texture generator then removes the tumor and regenerates tumors of different sizes in the same location. The tumor volume increases from left to right. In the upper right corner, the Pearson / Spearman correlation values between the texture features of the original and generated image are displayed. Breast: sagittal view, brain: axial view.}
        \label{fig6}
    \end{minipage}
    
    \centering
    \begin{minipage}{\textwidth}
        \vspace{3pt}
        \centering
        \includegraphics[width=0.825\textwidth]{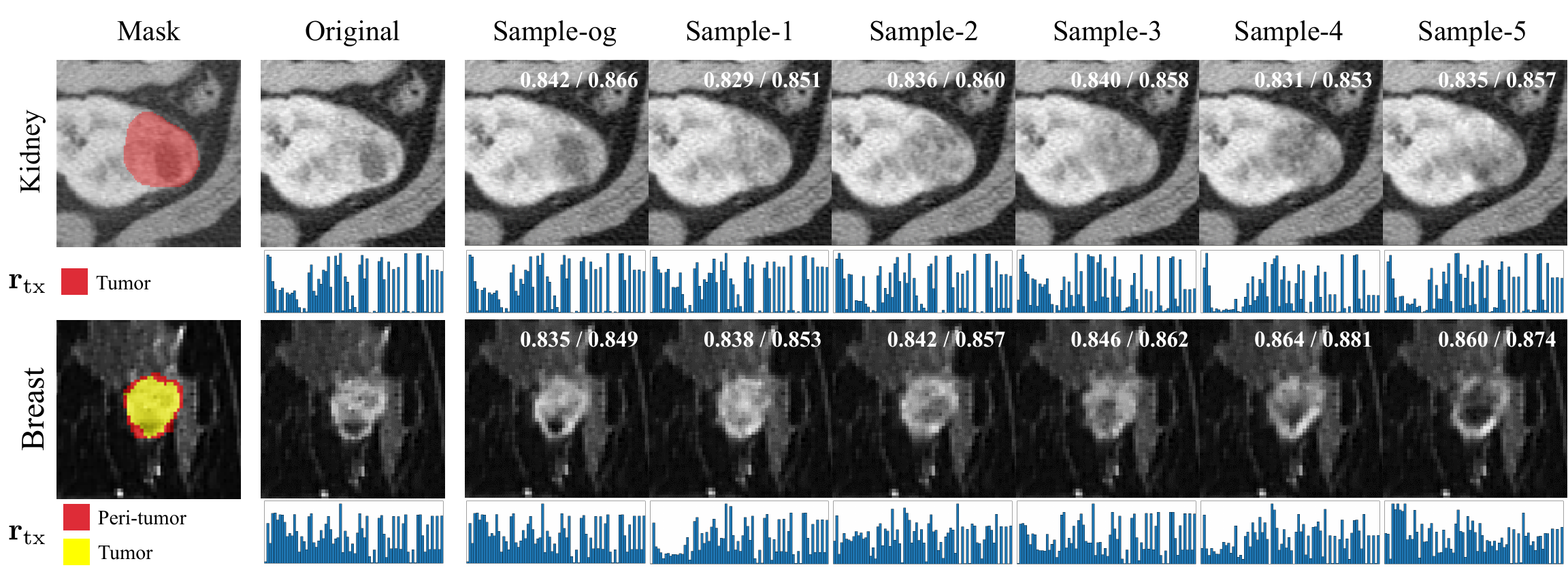}
        \vspace{-3pt}
        \caption{Results of tumor generation based on texture features. In the upper right corner, the Pearson / Spearman correlation values between the radiomics features given as the condition ($\mathbf{r}_\mathrm{tx}$) and the radiomics features extracted from the generated image are displayed. Sample-og denotes the generated images conditioned on the original radiomics features. A bar plot representing the various texture values is shown at the bottom of the image. For the kidney and breast, axial views are shown.}
        \label{fig7}
    \end{minipage}
    \vspace{-12pt}
\end{figure*}

We attempted to create tumors by adjusting the intuitive radiomics features of volume and sphericity. Figure \ref{fig5} shows the results of the tumors generated by manipulating these two features, presented through 3D rendering. The outcomes indicate that our shape generator has successfully reflected the shape features in creating the tumor masks. Therefore, this suggests that we can generate tumor masks of desired shapes by controlling the intuitive shape features.

In our study, we adjusted the volume size in the shape feature to generate tumor masks of different sizes. We then used these tumor masks to simulate changes based on tumor volume size. This experiment is depicted in Figure \ref{fig6}. Our proposed method not only generates tumors in a remarkably natural manner but also demonstrates the ability to simulate according to the volume size. Additionally, it shows that even as the volume size increases, the original texture is accurately reflected in the tumor generation as shown by correlation values over 0.8.


\noindent \textbf{Change Tumor Texture}: 
To verify if our proposed method accurately reflects texture features in tumor generation, we conducted experiments by generating tumors using texture features from different samples. That is, we replaced the original texture features of one subject with those of another subject. This experiment is depicted in Figure \ref{fig7}. To check how well the given texture is replicated, we extracted texture features from the generated image in the mask and compared their correlation with the features provided as conditions. The quantitative metrics shown in the figure indicate that the texture features were well-replicated in the generation process (correlation $>$ 0.8). This demonstrates that our model is capable of accurately reflecting even subtle textures that are not easily noticeable.

\begin{figure} [t]
    \vspace{-6pt}
    \centering
    \includegraphics[width=1\columnwidth]{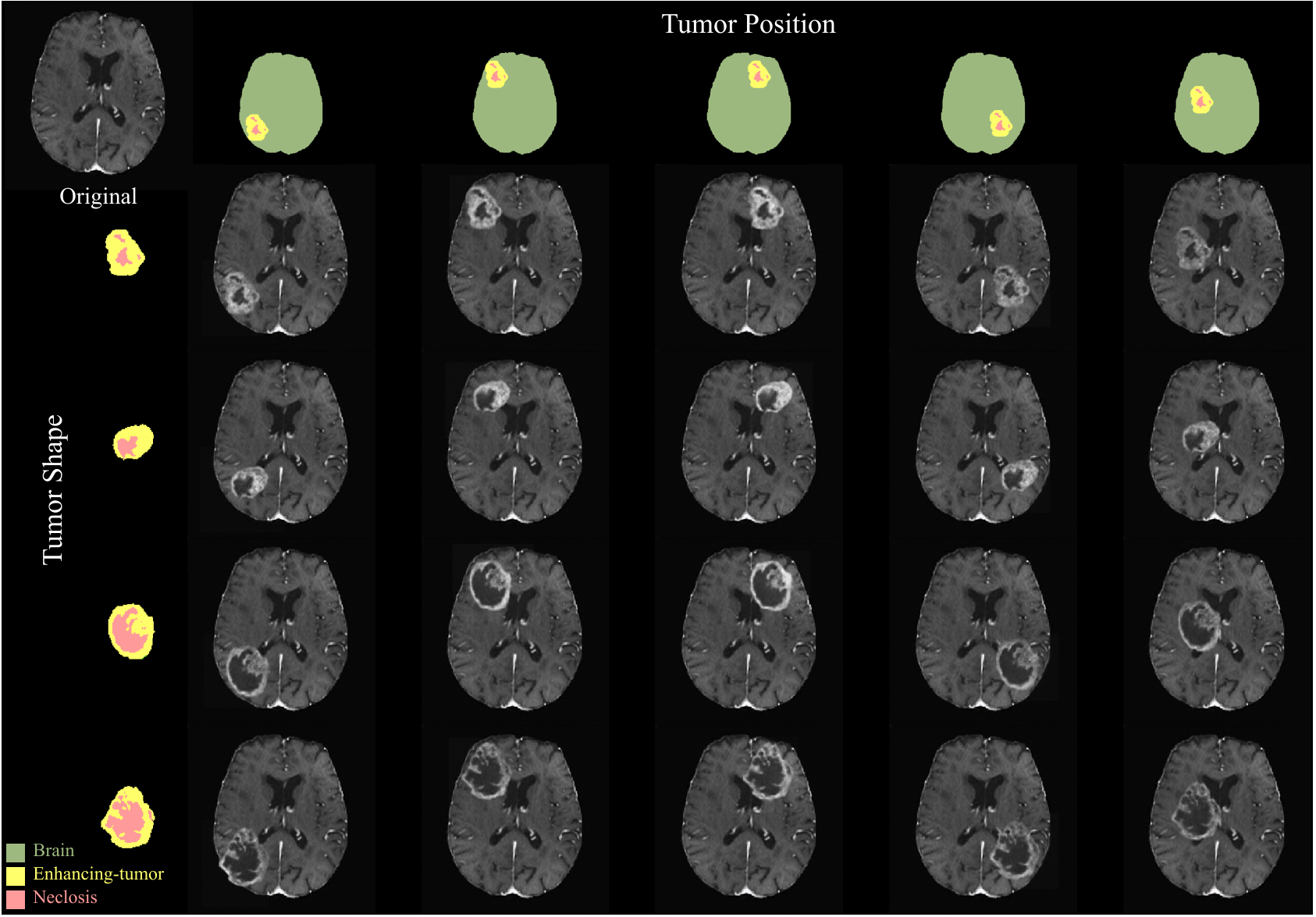}
    \vspace{-12pt}
    \caption{\jh{}{Results of tumor generation in the brain with various shapes and positions. The vertical axis represents different shapes, while the horizontal axis depicts various positions. The upper left indicates the original normal brain.}}
    \label{fig8}
\end{figure}

\noindent \textbf{Change Tumor Position}:
We conducted experiments on tumor generation by altering the shape and position of tumors, depicted in Figure \ref{fig8}. Tumors were generated in the normal brain without tumors and they were successfully created according to the specified shapes and positions. Our results show that our model can generate tumors of various shapes at different locations in an organ, without any constraints on position and shape. This highlights the potential for generating and utilizing a wider variety of samples in medical imaging, where data is often scarce.

\begin{table}[t]
    \centering
    \scalebox{0.75}{
        \begin{tabular}{lccc}
            \toprule
             Region &  \multicolumn{3}{c}{Brain Tumor} \\
            \cmidrule(lr){2-4} 
             Model \hspace{0.5cm} & ET & TC & All \\
            \midrule
            nnU-Net \cite{isensee2021nnu} & 0.9073\scriptsize{$\pm$0.176} \ & 0.7641\scriptsize{$\pm$0.301} \ & 0.8357\scriptsize{$\pm$0.256} \ \\
            nnU-Net (+Aug) & \ \textbf{0.9113\scriptsize{$\pm$0.166}}$^*$ & \ \textbf{0.7758\scriptsize{$\pm$0.294}}$^*$ & \ \textbf{0.8435\scriptsize{$\pm$0.256}}$^*$ \\
            \midrule
            Region & \multicolumn{3}{c}{Breast Tumor} \\
            \cmidrule(lr){2-4}
            Model \hspace{0.5cm} & Peri-Tumor & Tumor & All \\
            \midrule
            nnU-Net \cite{isensee2021nnu} & 0.8563\scriptsize{$\pm$0.043} \ & 0.8633\scriptsize{$\pm$0.048} \ & 0.8598\scriptsize{$\pm$0.046} \ \\
            nnU-Net (+Aug) & \ \textbf{0.8753\scriptsize{$\pm$0.031}}$^*$ & \ \textbf{0.8828\scriptsize{$\pm$0.032}}$^*$ & \ \textbf{0.8791\scriptsize{$\pm$0.031}}$^*$ \\
            \bottomrule
        \end{tabular}
    }
        \vspace{-6pt}
        \begin{flushright}
            \scalebox{1}{
                \footnotesize{$*p$-value$\ < 0.05$}
            }
        \end{flushright}
    \vspace{-12pt}
    \caption{Performance comparison between the baseline models and the model trained with additional synthesized images for downstream tasks of brain and breast tumor segmentation. Performance metrics include DICE scores.}
    \label{table4}
    \vspace{-9pt}
\end{table}

\noindent \textbf{Validation in a Downstream Task.}
We tested our generative model's effectiveness for downstream tasks by training a standard segmentation model with tumor-synthesized images to see if there was an improvement. The results are detailed in Table \ref{table4}. We trained the baseline segmentation model, nnU-Net \cite{isensee2021nnu}, and compared it across two tasks. In the brain tumor segmentation task, the baseline was trained with images from 1000 subjects, while another model was trained with an additional 1000 synthesized images for a total of 2000 images. In the breast tumor segmentation task, the baseline learned from 88 subject images, and a comparative model was trained with an additional 88 synthesized images, totaling 176 images. Brain tumor segmentation typically occurs in a multimodal setting. Since training was conducted solely with the T1ce modality, samples where the brain tumor is not clearly visible in T1ce present challenges for tumor detection, leading to a larger standard deviation in performance. The results showed a clear performance improvement when training with additional synthesized images, notably a larger boost in the less data-abundant breast tumor case than in brain tumors with more baseline data. This suggests that our method can significantly enhance model performance through augmentation, especially in situations with scarce or imbalanced datasets, highlighting its utility in augmenting models particularly when normal patient data outnumber abnormal cases.

\begin{table}
    \centering
    \vspace{-9pt}
    \begin{minipage}{\columnwidth}
        \centering
        \includegraphics[width=0.65\columnwidth]{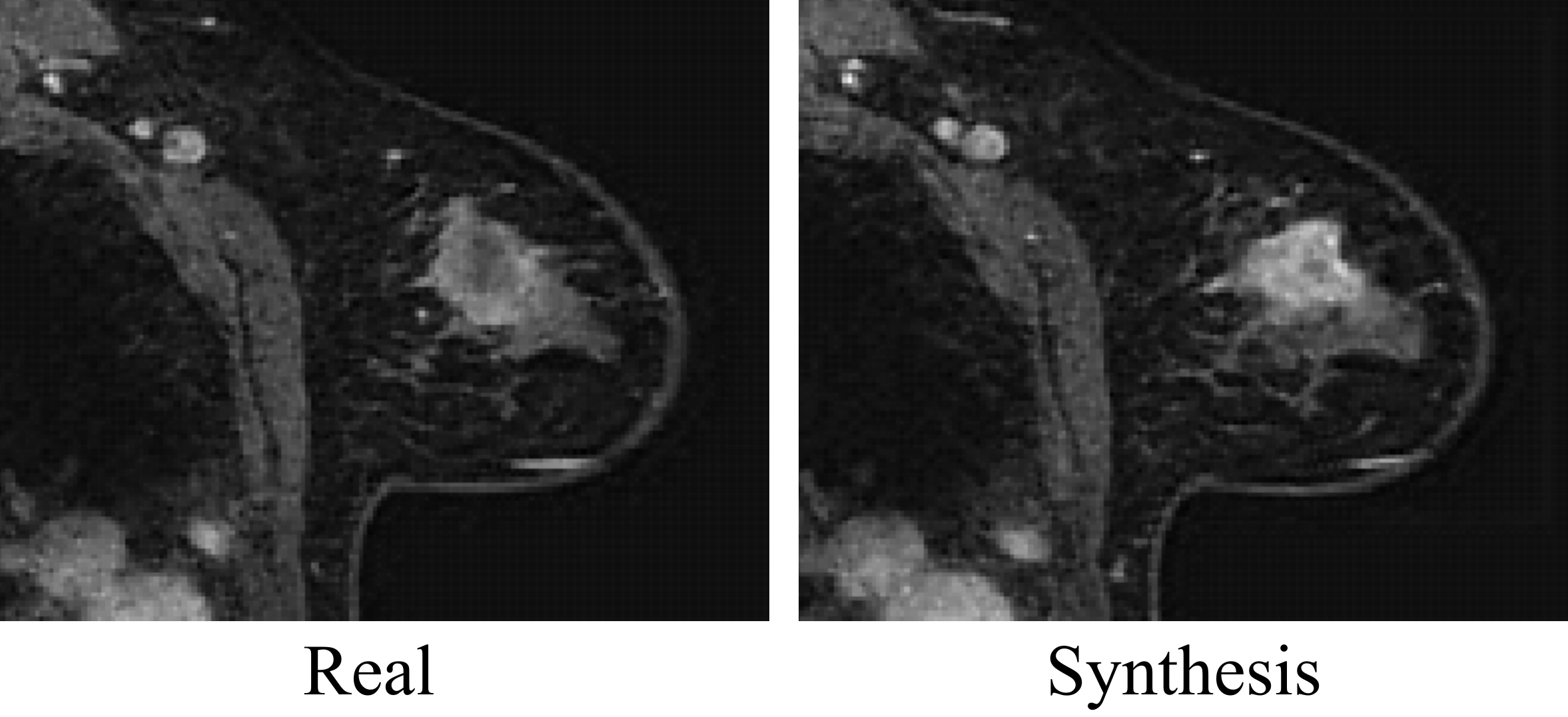}
        \vspace{-6pt}
        \captionsetup{type=figure}
        \caption{Illustration of real and synthesized images used for expert evaluation.}
        \label{fig10}
        \vspace{6pt}
    \end{minipage}
    \centering
    
    \begin{minipage}{\columnwidth}
        \centering
        \scalebox{0.75}{
            \begin{tabular}{cccc}
                \toprule
                Experts & \#1 & \#2 & Avg \\
                \cmidrule(lr){1-4}
                Real vs. Synthesis Acc & \quad 55\% \quad & \quad 60\% \quad & \quad 57.5\% \quad \\
                \bottomrule
            \end{tabular}
        }
    \end{minipage}
    \vspace{-6pt}
    \caption{Evaluation by two experts on real vs synthesized images.}
    \label{table5}
    \vspace{-9pt}
\end{table}

\noindent \textbf{Qualitative Evaluation by Experts.}
To verify the authenticity of synthesized images, we conducted expert evaluations by board-certified radiologists on 20 subjects. As depicted in Figure \ref{fig10}, experts were tasked with identifying real images from a mix of real and synthesized ones. This process assessed how convincingly the generated images mimic real ones. The results are detailed in Table \ref{table5}. These findings indicate that our model can produce images of such quality that even an expert would find it difficult to distinguish them from the real ones. 
Additional results of generating rare samples are in the Sec. \ref{sec:D_additional} of supplementary.
\section{Discussion}
\label{sec:discussion}

\noindent \textbf{Limitations}: In the case of tumors, there can be related changes in the tissues surrounding the tumor \cite{egeblad2010tumors, anderson2020tumor}. For instance, edema can occur around a brain tumor \cite{stummer2007mechanisms, bakas2017advancing}. Moreover, as a tumor grows, it also can push and distort the surrounding tissue \cite{mohammadi2018mechanisms}. In our experiment, we did not include such surrounding areas, which may limit the natural generation of tumors. Additionally, because we use radiomics features extracted from a single dataset for a given organ and modality, the learned distribution could be insufficient leading to degraded generalization on unseen data.



\noindent \textbf{Potential}: If it becomes possible to generate tumors of desired shapes and locations with desired textures, this would enable various simulations crucial for providing personalized treatment plans for tumor treatment. For instance, we could simulate scenarios of disease progression using simulated tumor images under an established prognosis model and possibly direct patients to alternative treatments. In short, our technology allows for simulations that assess risks based on the shape, texture, and position of the tumor.

\section{Conclusion}
\label{sec:conclusion}

In this study, we demonstrate the ability to synthesize 3D tumor images using a diffusion model conditioned on biologically grounded radiomics features. This generation capability not only enables appropriate data augmentation in medical imaging where data are scarce but also serves as a tool for simulating data for personalized treatment plans.

{\small
\bibliographystyle{ieee_fullname}
\bibliography{main}
}

\clearpage

\maketitle
\setlength{\abovedisplayskip}{\baselineskip}
\setlength{\belowdisplayskip}{\baselineskip}
\setcounter{page}{1}
\renewcommand{\thesection}{\Alph{section}}
\renewcommand{\thefigure}{\Alph{figure}}
\renewcommand{\thetable}{\Alph{table}}
\setcounter{section}{0}
\setcounter{figure}{0}
\setcounter{table}{0}

This supplementary material provides additional details not included in the main paper due to space constraints.
In \cref{sec:A_datasets}, we provide detaileds about the four datasets we employed.
\cref{sec:B_implementation} includes information on training details and network hyperparameters.
\cref{sec:C_radiomics} details the features used as conditions in our model.
\cref{sec:D_additional} represents additional results for the rare sample generation.
\cref{sec:E_future} describes future works.
Finally, \cref{sec:F_training} details the training and inference algorithm.

\section{Datasets} 
\label{sec:A_datasets}

\noindent \textbf{Kidney}: We utilize the 2023 Kidney and Kidney Tumor Segmentation Challenge (KiTS23) dataset \cite{heller2023kits21}. KiTS23 is a challenge dataset designed for segmenting kidneys and kidney tumors in CT scans. In our study, we used 406 tumor subjects out of 489 in the provided challenge training dataset, splitting them into 325 for training and 81 for the test set. \jh{}{During training, the 3D patch volume was cropped from the center of the tumor mask and patch volume size of $112\times112\times96$ was employed}. We clipped the intensities to a range of [-175, 250] Hounsfield unit (HU) and further normalized them to [0,1].

\vspace{1mm}

\noindent \textbf{Lung}: We employ the non-small cell lung cancer (NSCLC) dataset \cite{aerts2015data}. This dataset consists of CT scans. In our study, we utilized 417 tumor subjects out of the provided training dataset of 422, dividing them into 334 for training and 83 for the test set. \jh{}{During training, the 3D patch volume was cropped from the center of the tumor mask and volume size of $112\times112\times80$ was used}. The HU range of [-1000, 1000] was applied and normalized to [0,1].

\vspace{1mm}

\noindent \textbf{Breast}: We utilize a private dataset, which is a Dynamic Contrast Enhancement MRI (DCE-MRI) dataset. MRI scans were conducted using either a 1.5-T or a 3.0-T scanner from Philips. The scans included axial imaging with one pre-contrast and six post-contrast dynamic series. Contrast-enhanced images were acquired at 0.5, 1.5, 2.5, 3.5, 4.5, and 5.5 minutes after the contrast injection. We used the images from 0.5 minutes after contrast injection. The dataset comprises 110 breast cancer patients, of which 88 were used for the training set and 22 for the test set. \jh{}{During training, the 3D patch volume was cropped from the center of the tumor mask and patch volume size of $112\times112\times96$ with nonzero normalization (normalization excluding zero values, see Monai \footnote{\url{https://monai.io/}} framework) for MRI images was employed.}

\vspace{1mm}

\noindent \textbf{Brain}: We utilize The Brain Tumor Segmentation Challenge 2021 (BraTS 2021) dataset \cite{baid2021rsna, menze2014multimodal, bakas2017advancing}, which includes multimodal MRI scans such as T1-weighted (T1), contrast-enhanced T1-weighted (T1ce), T2-weighted (T2), and Fluid Attenuated Inversion Recovery (FLAIR). Among these, T1ce, a modality enhanced with contrast agents to better visualize tumors, was employed in our study. From the BraTS 2021 training set, we utilized data from 1204 out of 1251 subjects, allocating 1000 for training and 204 for testing. \jh{}{During training, 3D patch volumes were centered around the enhancing-tumor mask, with a patch volume size of $112\times112\times96$ with nonzero normalization was applied to the MRI images.}

\section{Implementation Details} 
\label{sec:B_implementation}

\subsection{Training Details}
The network was trained using the Adam \cite{kingma2014adam} optimizer with a learning rate set to $5\times10^{-6}$. The training was conducted on four A100 80GB GPUs with a batch size of one per GPU. The model was built using PyTorch version 1.13.1.

\subsection{Comparison Methods}

\subsubsection{GAN-based Model.} The GAN-based model is adapted by modifying the 3D image translation model Ea-GAN \cite{yu2019ea} (based on pix2pix \cite{isola2017image}), with the addition of cross-attention. The model architecture, based on the UNet \cite{ronneberger2015u} structure, is specified in Table \ref{tableA}.

\begin{table}[h]
    \centering
    \resizebox{1\columnwidth}{!}{%
        \begin{tabular}{ccccclccc}
            \toprule
             Stream & \multicolumn{3}{c}{Cross Attn}& \multicolumn{2}{c}{Act.}& Conv. & Norm. & Out ch.\\
             \midrule
             \textbf{In}& \multicolumn{3}{c}{}&  \multicolumn{2}{c}{}& $C$ &  & 64\\
             \midrule
             \textbf{DownBlock}& \multicolumn{3}{c}{}& \multicolumn{2}{c}{LeakyReLU}& $C$ & IN & \text{[}128,256\text{]}\\
            \midrule
             \textbf{DownBlock}& \multicolumn{3}{c}{\checkmark }& \multicolumn{2}{c}{LeakyReLU}& $C$& IN& \text{[}256,512\text{]}\\
             \midrule
             \textbf{UPBlock}& \multicolumn{3}{c}{\checkmark }&  \multicolumn{2}{c}{ReLU}&  $C^T$&  IN& \text{[}512, 256\text{]}\\
            \midrule
            \textbf{UPBlock}& \multicolumn{3}{c}{}& \multicolumn{2}{c}{ReLU}& $C^T$& IN&\text{[}256,128\text{]}\\ 
            \midrule
            \textbf{Out}& \multicolumn{3}{c}{}& \multicolumn{2}{c}{ReLU}& $C^T$& Tanh&1\\
            \bottomrule
        \end{tabular}
        }
    \vspace{6pt}
    \caption{Details of GAN-based Model. $C$ is the convolution layer and $C^T$ is the convolution transpose layer with $4\times4\times4$ kernel, $2\times2\times2$ stride, and $1\times1\times1$ padding. IN is the instance normalization layer. The Out layer uses tanh as the activation function to generate the final output.} 
    \label{tableA}
\end{table}

\subsubsection{Latent Diffusion Model.} LDM \cite{rombach2022high} is adapted to a 3D format from its original method for comparison. LDM consists of two components: a pretrained VQGAN \cite{esser2021taming} and a DDPM \cite{ho2020denoising}. The detailed structure of this model is outlined in Table \ref{tableB}. During training, the number of time steps for the diffusion process was set to 1000. In the inference stage, the DDIM \cite{song2020denoising} method was employed, utilizing 200 sampling steps.

\begin{table*}[h!]
    \centering
    \caption{Model structural details of LDM and BBDM used for the tumor texture generation task. $|\mathcal{Z}|$ represents codebook size in the latent space.} 
    \vspace{-6pt}
    \resizebox{\textwidth}{!}{%
        \begin{tabular}{lccccccc}
            \toprule
             Model & $z$-shape & $|\mathcal{Z}|$ & \quad Training Steps \quad & \quad Noise schedule \quad & \quad Channel Multiplier \quad &  \quad Channels \quad & \quad Model Size \quad \\
             \midrule
             LDM-f2        & $56\times56\times48\times2$ & \quad 2048 \quad & 1000 & linear & [1,4,8] & 128 & 658.32M \\
             BBDM-f2 \quad & $56\times56\times48\times2$ & \quad 2048 \quad & 1000 & linear & [1,4,8] & 128 & 681.00M\\
            \bottomrule
        \end{tabular}
        }
    \label{tableB}
\end{table*}

\begin{table*}[h!]
    \centering
    \caption{Model structure details of the GigaGAN used for the tumor shape generation task.}
    \vspace{-6pt}
    \resizebox{\textwidth}{!}{%
        \begin{tabular}{lccccccc}
            \toprule
             Model & \quad $\mathbf{z}$ dim \quad & \quad $\mathbf{w}$ dim \quad & \quad $G$ Channels \quad & \quad $D$ Channels \quad & \quad $G$ Attention Resolution \quad & \quad $D$ Attention Resolution \quad & \quad Attention Type \quad \\
             \midrule
             Shape 24                & 128 & 512 & 512 & 512 & [6,12] & [6,12] & self + cross \\
             Shape 24$\rightarrow$96 & 128 & 512 & 512 & 512 & [12,24] & [12,24] & self + cross \\
            \bottomrule
        \end{tabular}
        }
    \label{tableC}
\end{table*}

\begin{table*}[h!]
    \centering
    \caption{Hyperparameters of the Exponential Moving Average (EMA) and ReduceLROnPlateau learning rate scheduler used in the training process.} 
    \vspace{-6pt}
    \resizebox{\textwidth}{!}{%
        \begin{tabular}{lcccccccccc}
            \toprule
             & \multicolumn{3}{c}{EMA Parameters} & & \multicolumn{6}{c}{LR Scheduler Parameters} \\
             \midrule
             Model & \quad Start Step \quad & \quad Decay \quad & \quad Update Interval \quad & & \quad Max lr \quad & \quad Min lr \quad & \quad Factor \quad & \quad patience \quad & \quad Cool Down \quad & \quad Threshold \quad\\
             \midrule
             BBDM-f2 & 30000 & 0.995 & 8 & & 5.0e-6 & 5.0e-7 & 0.5 & 3000 & 3000 & 1.0e-4 \\
            \bottomrule
        \end{tabular}
        }
    
    \label{tableD}
\end{table*}

\subsection{Network hyperparameters}  

\subsubsection{Tumor Shape Generator.} The Tumor Shape Generator is based on GigaGAN \cite{kang2023scaling}, which has shown successful results in generating images from text in 2D natural images. This generator adapts GigaGAN to a 3D format and uses a shape feature-to-image approach to create tumor masks. Additionally, the Tumor Shape Generator employs cascaded generation processes where images are initially generated at a resolution of $24\times24\times24$ and then upsampled to $96\times96\times96$. Detailed information about its structure can be found in Table \ref{tableC}.

\vspace{1mm}

\subsubsection{Tumor Texture Generator.} The Tumor Texture Generator is based on BBDM \cite{li2023bbdm} and has been modified for 3D application. This generator comprises two stages: a pretrained VQGAN and a BBDM. The detailed structure of the model is specified in Table \ref{tableB}. During the training phase, the number of time steps for the Brownian Bridge was established at 1000, whereas, in the inference stage, 200 sampling steps were utilized. The training parameters for BBDM are specified in Table \ref{tableD}.

\begin{table*}[h!]
    \centering
    \caption{\textbf{Radiomics features used in our study.} There are 16 shape, 18 histogram, 24 GLCM, 16 GLSZM, and 16 GLRLM features. The tumor shape generator utilizes 16 shape features for each region of interest (ROI) type, while the tumor texture generator employs 74 texture features for each ROI type.} 
    \vspace{-6pt}
    \resizebox{\textwidth}{!}{%
        \begin{tabular}{c||c||c||c||c}
            \toprule
             \textbf{Shape} & \textbf{Histogram} & \textbf{GLCM} & \textbf{GLSZM} & \textbf{GLRLM} \\
             Mesh Volume & Energy & Autocorrelation& Small Area Emphasis& Short Run Emphasis\\
             Surface Area & Entropy & Joint Average& Large Area Emphasis& Long Run Emphasis\\
             Surface Area to Volume ratio & Minimum  & Cluster Prominence& Gray Level Non-Uniformity& Gray Level Non-Uniformity\\
             Sphericity & Maximum  & Cluster Shade& Gray Level Non-Uniformity Normalized& Gray Level Non-Uniformity Normalized\\
             Compactness 1 & Mean& Cluster Tendency& Size-Zone Non-Uniformity& Run Length Non-Uniformity\\
             Compactness 2 & Median & Contrast& Size-Zone Non-Uniformity Normalized& Run Length Non-Uniformity Normalized\\
             Spherical Disproportion & 10th percentile   & Correlation& Zone Percentage& Run Percentage\\
             Maximum 3D diameter & 90th percentile  & Difference Average& Gray Level Variance& Gray Level Variance\\
             Maximum 2D diameter (Slice) & Interquartile Range   & Difference Entropy& Zone Variance& Run Variance\\
             Maximum 2D diameter (Column) & Range   & Difference Variance& Zone Entropy& Run Entropy\\
             Maximum 2D diameter (Row) & Mean Absolute Deviation  & Joint Energy& Low Gray Level Zone Emphasis& Low Gray Level Run Emphasis\\
             Major Axis Length & Robust Mean Absolute Deviation  & Joint Entropy& High Gray Level Zone Emphasis& High Gray Level Run Emphasis\\
             Minor Axis Length & Root Mean Squared  & Informational Measure of Correlation 1 & Small Area Low Gray Level Emphasis& Short Run Low Gray Level Emphasis\\
             Least Axis Length & Standard Deviation  & Informational Measure of Correlation 2 & Small Area High Gray Level Emphasis& Short Run High Gray Level Emphasis\\
             Elongation & Skewness & Inverse Difference Moment  & Large Area Low Gray Level Emphasis& Long Run Low Gray Level Emphasis\\
             Flatness & Kurtosis &  Maximal Correlation Coefficient   & Large Area High Gray Level Emphasis& Long Run High Gray Level Emphasis\\
             & Variance & Inverse Difference & & \\
             & Uniformity & Inverse Difference Normalized & & \\
             & & Inverse Difference Moment Normalized & & \\
             & & Inverse Variance  & & \\
             & & Maximum Probability  & & \\
             & & Sum Average  & & \\
             & & Sum Entropy  & & \\
             & & Sum of Squares & & \\
            \bottomrule
        \end{tabular}
        }
    \label{tableE}
\end{table*}

\section{Radiomics Features} 
\label{sec:C_radiomics}
\jh{}{
In our study, we extracted radiomics \cite{aerts2014decoding} features using PyRadiomics \footnote{\url{https://pyradiomics.readthedocs.io/}} \cite{van2017computational}. We obtained shape, histogram, and texture features for model training, and the details of each feature are described in the subsection. 
Additionally, most feature definitions used in this study are provided in a Zwanenburg et al \cite{zwanenburg2020image}. 
The features used are specified in Table \ref{tableE}. Additionally, Figure \ref{figA} depicts the correlation matrix of each feature to illustrate the relationships between them.
}
\subsection{Shape feature} 
For our Tumor Shape Generation task, we utilized a total of 16 shape features  for each type of ROI. Shape features quantitatively represent the physical form and structure of the ROI. These morphological characteristics, such as the size, shape, and orientation of tumors, are analyzed for diagnosing diseases and evaluating prognoses.

The main shape features include:
\begin{itemize}
\item[•] Volume: Measures the overall volume of the tumor.
\item[•] Surface Area: Measures the surface area of the tumor.
\item[•] Sphericity: Indicates the ratio of the tumor's length and width and thus reflects how close the tumor is to being spherical.
\end{itemize}

During the experiment to manipulate tumor shape, we focused on changing the volume and sphericity. Volume is just the sum of voxels in mm. Sphericity is made of two components, volume and surface area. Thus, we manipulated volume and surface area to adjust the two features. Other shaped features that depended on volume and surface area were adjusted accordingly.

\subsection{Texture feature} 
For our tumor texture generation task, we utilized a total of 74 features for each type of ROI, which included 18 histogram features, 24 GLCM features, 16 GLSZM features, and 16 GLRLM features. Detailed descriptions of each feature are provided in the subsections below.

\vspace{1mm}
\noindent \textbf{Histogram}: Histogram features, also known as first-order features, are derived from the distribution of pixel or voxel intensities within the ROI. These features provide valuable insights into the texture of the tissue by analyzing the intensity histogram of the ROI.

The main histogram features include:
\begin{itemize}
\item[•] Median: The middle-intensity value when all values are sorted.
\item[•] Skewness: The asymmetry in the intensity distribution.
\item[•] Energy: The sum of squared intensities, representing the magnitude of voxel values.
\item[•] Entropy: The randomness or complexity of the intensity distribution.
\end{itemize}

\vspace{1mm}
\noindent \textbf{Gray Level Co-occurrence Matrix (GLCM)}: GLCM is a method used in image processing to examine the texture of an image by assessing how often pairs of pixels with specific values and in a specified spatial relationship occur in an image, creating a GLCM, and then extracting statistical measures from this matrix.

The main GLCM features include:
\begin{itemize}
\item[•] Contrast: Measures the local variations in the GLCM.
\item[•] Correlation: Assesses how correlated a pixel is to its neighbors.
\item[•] Inverse Difference Moment (or Homogeneity): Measures the closeness of the distribution of elements in the GLCM to the GLCM diagonal.

\end{itemize}

\vspace{1mm}
\noindent \textbf{Gray Level Size Zone Matrix (GLSZM)}: 
GLSZM is a method used in radiomics for texture analysis, particularly focusing on the size and distribution of continuous zones with the same gray level intensity in an image. GLSZM provides a way to quantify patterns and structures in an image that are not captured by first-order statistics or other texture matrices like the GLCM. A zone in GLSZM refers to a group of connected pixels that have the same gray level intensity. 

The main GLSZM features include:
\begin{itemize}
\item[•] Small Area Emphasis: Focuses on the distribution of small size zones.
\item[•] Large Area Emphasis: Highlights the presence of large size zones.
\item[•] Zone Percentage: The total number of zones relative to the size of the ROI, indicating textural uniformity.
\item[•] Low Gray Level Zone Emphasis: Reflects the proportion of zones with lower gray level values.
\item[•] High Gray Level Zone Emphasis: Indicates the presence of zones with higher gray level values.
\end{itemize}

\vspace{1mm}
\noindent \textbf{Gray Level Run Length Matrix (GLRLM)}: 
GLRLM focuses on examining the length and directionality of continuous runs of pixels with the same gray level intensity in an image, thus providing important information about the texture and structural patterns. A run in GLRLM is defined as a set of consecutive, collinear pixels having the same gray level intensity. The length of a run is the number of pixels in this set.

The main GLRLM features include:
\begin{itemize}
\item[•] Short Run Emphasis: Measures the distribution of short runs, indicating fine textural patterns.
\item[•] Long Run Emphasis: Highlights long runs, suggesting coarser textures.
\item[•] Run Length Nonuniformity: Quantifies the variability of run lengths, with higher values indicating more heterogeneous textures.
\item[•] Gray Level nuniformity: Measures the variability of gray levels in the runs.
\end{itemize}

\section{Additional Results}
\label{sec:D_additional}
\begin{figure}[h]
    \centering
    \includegraphics[width=\columnwidth]{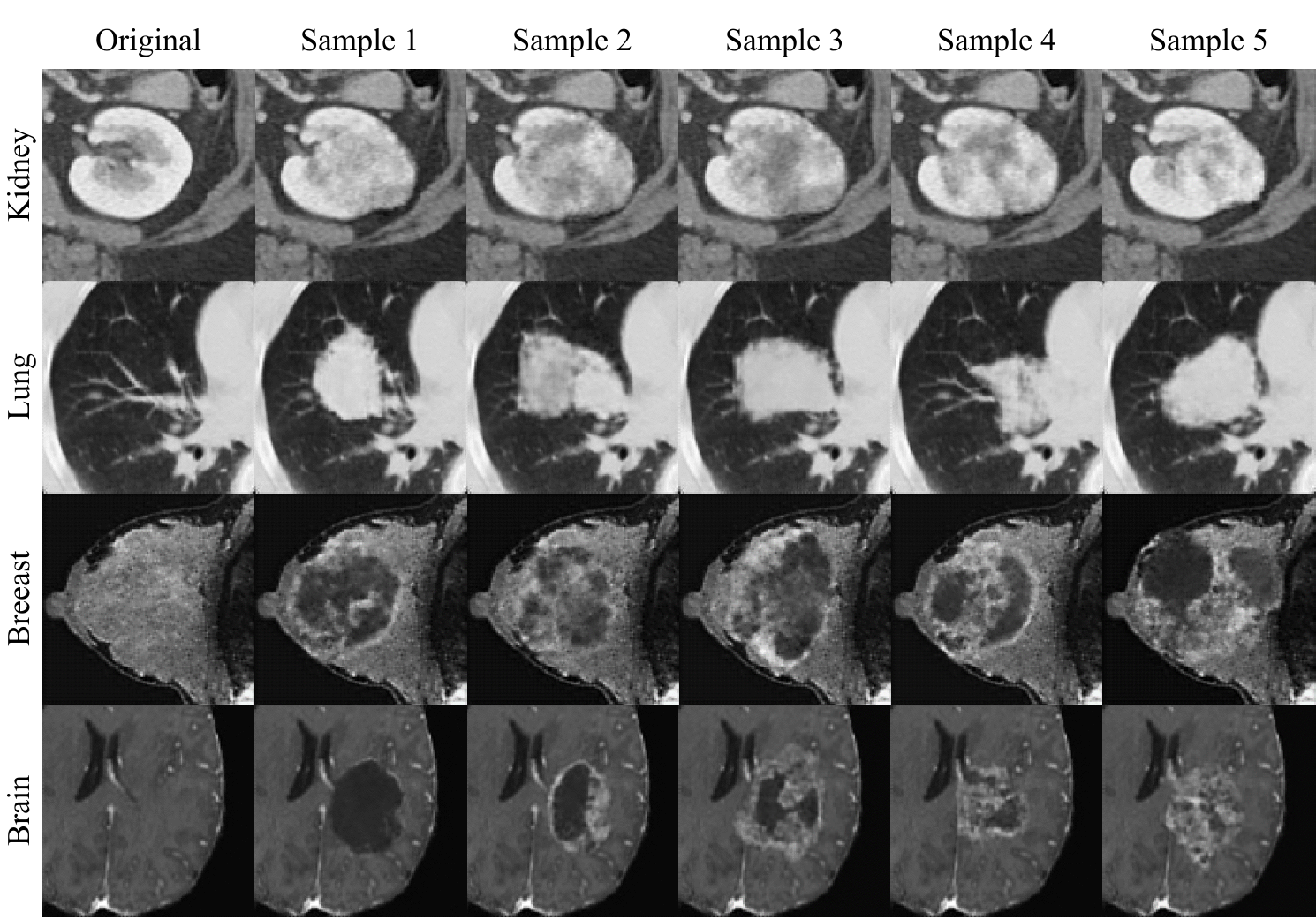}
    \caption{Results of generating rare samples. }
    \label{fig9}
\end{figure}

\begin{figure*} [t]
    \centering
    \includegraphics[width=0.95\textwidth]{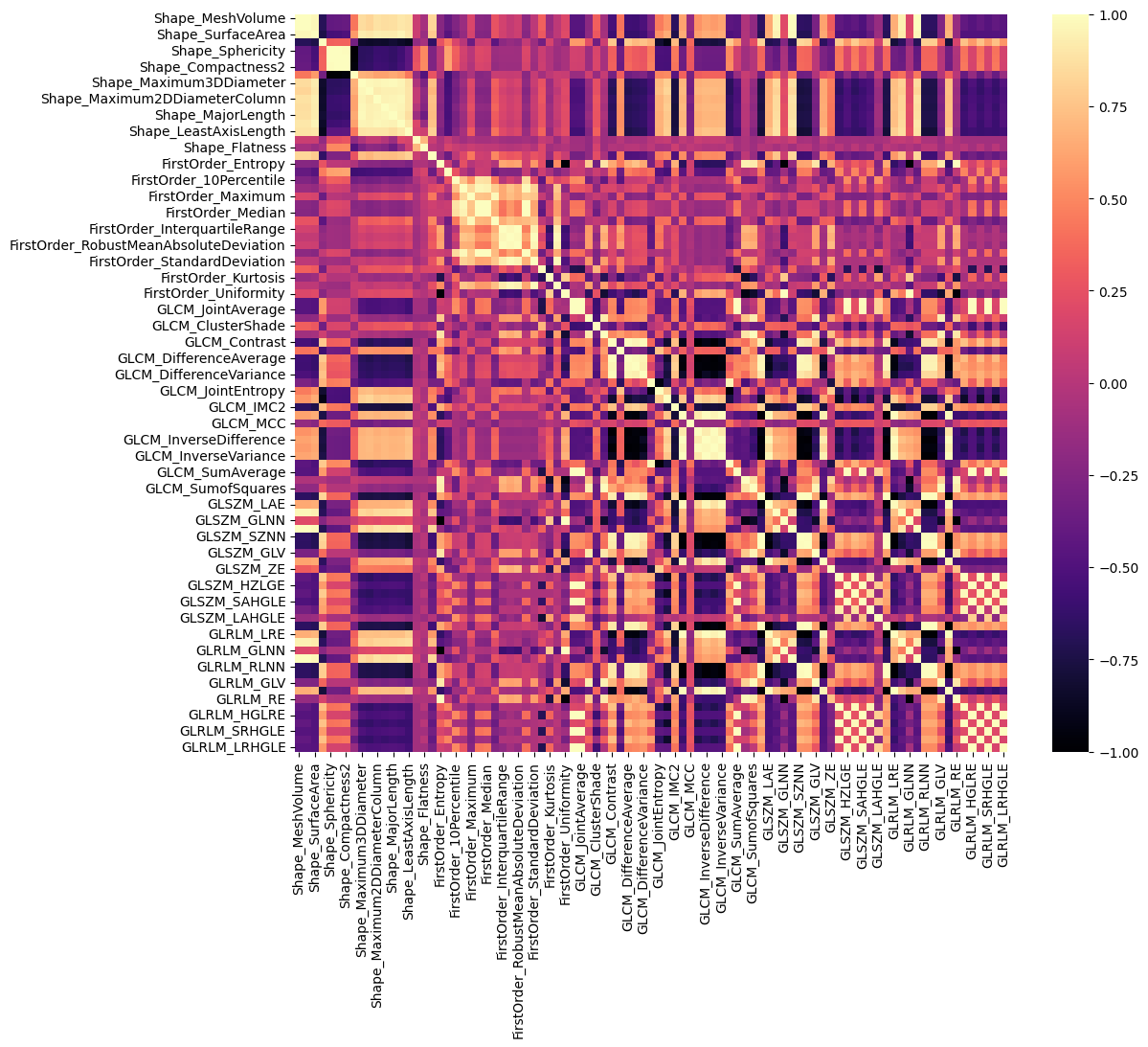}
    \caption{\textbf{Radiomcis feature correlation matrix} extracted from the tumor region of a breast MRI. A higher positive correlation between features approaches 1, a higher negative correlation approaches -1, and a lower correlation approaches 0.}
    \label{figA}
\end{figure*}

To demonstrate the effectiveness of our proposed method in simulations, we attempted to generate large tumors, which are rare in clinical settings, This process is depicted in Figure \ref{fig9}. Cases with such large tumors are unusual in medical imaging and difficult to find data. Our results prove that we can generate these rare samples. Furthermore, not only can we create hard-to-find samples, but we can also apply our method to simulate the growth of tumors. We anticipate this will contribute to prognostic research studies.

\section{Future works}
\label{sec:E_future}
Medical imaging tends to be sensitive to parameter variations across different acquisition sites, which could lead to potential inconsistencies in extracting intensity-sensitive radiomics features, because minor alterations may result in significantly different feature values. This presents a challenge when attempting to apply the process across diverse domains and datasets simultaneously. For example, difficulties arise not only between CT and MRI but also across datasets from different anatomical organs like breast and brain. Advancements in research, in tandem with domain normalization for extracting and standardized radiomics features within a unified space, could pave the way for substantial progress.

\section{Training and Inference Process}
\label{sec:F_training}
The training and inference processes of the texture generator involve scenarios where texture features are either provided (for generating tumors) or not provided (for generating normal tissue). During the training of non-tumor regions, an area that does not overlap with the tumor region is randomly chosen for masking. Subsequently, this masked area is reconstructed and utilized in training. These steps are summarized in Algorithm \ref{algorithm1}.

For sampling a normal image, the intended area for restoration to its normal state is masked, followed by a diffusion process to derive the normal image. This method is detailed in Algorithm \ref{algorithm2}. When sampling a tumor image, a mask is generated with specific shape features using a shape generator. This mask is applied to the intended area for tumor generation. A diffusion process is then carried out, conditioned on the texture features, to produce the tumor image. This method is detailed in Algorithm \ref{algorithm3}.

\clearpage 

\clearpage

 \begin{algorithm*} [h]
    \scalebox{1}{
    \begin{minipage}{1\linewidth}
        \caption{Texture generator training loop with image $I$, mask $M$, radiomics texture feature $\mathbf{r}_{\mathrm{tx}}$, and VQGAN encoder $\mathcal{E}$}
        \begin{algorithmic}[1]
        \Repeat
            \If{$\text{Training for Tumor}$}
                \State $I^M \leftarrow I \odot M$ \Comment{Masking image with tumor mask}
            \Else
                \State $M' \leftarrow \text{Shift}(M)$ \Comment{Shift mask position without overlapping tumor}
                \State $I^M \leftarrow I \odot M'$ \Comment{Masking image with tumor mask}
                \State $\mathbf{r}_{\mathrm{tx}} \leftarrow \text{None}$ \Comment{Initialize radiomics texture feature to none}
            \EndIf
            \State $\boldsymbol{x}_0=\mathcal{E}(I), \boldsymbol{y}=\mathcal{E}(I^M)$ \Comment{Image compression with VQGAN encoder}
            \State $\boldsymbol{x}_0 \sim q(\boldsymbol{x}_0), \boldsymbol{y} \sim q(\boldsymbol{y})$
            \State $\text{timestep} \, t \sim Uniform(1,...,T)$
            \State $\text{Gaussian noise} \, \epsilon \sim \mathcal{N}(0,\mathbf{I})$
            \State $\text{Forward diffusion} \, \boldsymbol{x}_t = (1-m_t)\boldsymbol{x}_0 + m_t\boldsymbol{y} + \sqrt{\delta_t}\epsilon$
            \State $\text{Take gradient descent step on}$ \\
            $\hspace{12mm} \bigtriangledown_\theta||m_t(\boldsymbol{y} - \boldsymbol{x}_0) + \sqrt{\delta_t}\epsilon - \epsilon_\theta(\boldsymbol{x}_t, t, \mathbf{r}_{\mathrm{tx}})||^2$
        \Until{converged}
        \end{algorithmic}
        \label{algorithm1}
    \end{minipage}
    }
\end{algorithm*}

 \begin{algorithm*} [h]
\caption{Normal image sampling with VQGAN decoder $\mathcal{D}$}
\scalebox{1}{
\begin{minipage}{1\linewidth}
\begin{algorithmic}[1]
\State $I^M \leftarrow I \odot M$ \Comment{Masking image with tumor mask}
\State $\boldsymbol{y} = \mathcal{E}(I^M)$ \Comment{Image compression with VQGAN encoder} 
\State $\boldsymbol{x}_T=\boldsymbol{y} \sim q(\boldsymbol{y})$ \Comment{Sample conditional input}
\For{$t=T,...,1$}
    \State $\textbf{z} \sim \mathcal{N}(0, \textbf{I}) \; \text{if} \; t > 1, \text{else} \; \textbf{z} = 0$
    \State $\boldsymbol{x}_{t-1} = \boldsymbol{c}_{xt}\boldsymbol{x}_{t} + \boldsymbol{c}_{yt}\boldsymbol{y}-\boldsymbol{c}_{\epsilon t}\epsilon_{\theta}(\boldsymbol{x}_{t}, t) + \sqrt{\tilde{\delta_t}}\boldsymbol{z}$
\EndFor
\State \textbf{return} $\mathcal{D}(\boldsymbol{x}_0)$ \Comment{Reconstruction with VQGAN decoder}
\end{algorithmic}
\label{algorithm2}
\end{minipage}
}
\end{algorithm*}

 \begin{algorithm*} [h]
\caption{Tumor image sampling with shape generator G and radiomics shape feature $\mathbf{r}_{\mathrm{sh}}$}
\scalebox{1}{
\begin{minipage}{1\linewidth}
\begin{algorithmic}[1]
\State $ z \sim \mathcal{N}(0,\textbf{I})$
\State $ \mathcal{M} = G(z, \mathbf{r}_{\mathrm{sh}})$ \Comment{Generate tumor mask with shape feature}
\State $I^\mathcal{M} \leftarrow I \odot \mathcal{M}$ \Comment{Masking image with tumor mask}
\State $\boldsymbol{y} = \mathcal{E}(I^\mathcal{M})$  \Comment{Image compression with VQGAN encoder}
\State $\boldsymbol{x}_T=\boldsymbol{y} \sim q(\boldsymbol{y})$ \Comment{Sample conditional input}
\For{$t=T,...,1$}
    \State $\textbf{z} \sim \mathcal{N}(0, \textbf{I}) \; \text{if} \; t > 1, \text{else} \; \textbf{z} = 0$
    \State $\boldsymbol{x}_{t-1} = \boldsymbol{c}_{xt}\boldsymbol{x}_{t} + \boldsymbol{c}_{yt}\boldsymbol{y}-\boldsymbol{c}_{\epsilon t}\epsilon_{\theta}(\boldsymbol{x}_{t}, t, \mathbf{r}_{\mathrm{tx}}) + \sqrt{\tilde{\delta_t}}\boldsymbol{z}$
\EndFor
\State \textbf{return} $\mathcal{D}(\boldsymbol{x}_0)$  \Comment{Reconstruction with VQGAN decoder}
\end{algorithmic}
\label{algorithm3}
\end{minipage}
}
\end{algorithm*}

\clearpage 

\end{document}